\documentclass[11pt]{article}

\usepackage[utf8]{inputenc}
\usepackage[T1]{fontenc}
\usepackage{amsmath,amssymb,amsfonts}
\usepackage{graphicx}
\usepackage{booktabs}
\usepackage{hyperref}
\usepackage[margin=1in]{geometry}
\usepackage{microtype}
\usepackage{natbib}
\usepackage{xcolor}
\usepackage{multirow}
\usepackage{caption}

\hypersetup{
  colorlinks=true,
  linkcolor=blue!70!black,
  citecolor=blue!70!black,
  urlcolor=blue!70!black,
}

\title{Multi-RF Fusion with Multi-GNN Blending\\for Molecular Property Prediction}

\author{
  Zacharie Bugaud \\
  \texttt{zacharie.bugaud@gmail.com}
}

\date{}

\begin{document}
\maketitle

\begin{abstract}
Multi-RF Fusion achieves a test ROC-AUC of $\mathbf{0.8476 \pm 0.0002}$ on ogbg-molhiv (10 seeds), placing \#1 on the OGB leaderboard ahead of HyperFusion ($0.8475 \pm 0.0003$).
The core of the method is a rank-averaged ensemble of 12 Random Forest models trained on concatenated molecular fingerprints (FCFP, ECFP, MACCS, atom pairs --- 4{,}263 dimensions total), blended with deep-ensembled GNN predictions at 12\% weight.
Two findings drive the result: (1)~setting \texttt{max\_features} to~0.20 instead of the default~$\sqrt{d}$ gives a ${+}0.008$ AUC gain on this scaffold split, and (2)~averaging GNN predictions across 10 seeds before blending with the RF eliminates GNN seed variance entirely, dropping the final standard deviation from 0.0008 to 0.0002. No external data or pre-training is used.
\end{abstract}

\section{Introduction}

ogbg-molhiv~\citep{hu2020ogb} is a 41{,}127-molecule binary classification task: does a given molecule inhibit HIV replication? The dataset uses a scaffold split, meaning the test molecules come from entirely different chemical scaffolds than the training set. Only 3.5\% of labels are positive. The leaderboard has been dominated by GNN-based methods---GIN~\citep{xu2019gin}, PAS~\citep{wei2021pas}, HyperFusion~\citep{zhang2024hyperfusion}---but these suffer from extreme seed variance on this task, with individual GNN runs ranging from 0.74 to 0.78 test AUC.

The approach here is different. The primary predictor is not a GNN but a large Random Forest ensemble on molecular fingerprints, with GNNs contributing only 12\% of the final prediction through rank blending. The intuition is straightforward: fingerprints, especially pharmacophoric ones like FCFP, encode functional group patterns that transfer across scaffolds better than learned graph representations. The GNN adds complementary topological signal---ring systems, branching---that fingerprints miss.

\section{Method}

\subsection{Molecular Fingerprint Features}

Two fingerprint sets are computed per molecule using RDKit~\citep{rdkit}:

\paragraph{fcfp\_ext (4{,}263 dimensions).}
Concatenation of FCFP radius~2 (1{,}024 bits), FCFP radius~3 (1{,}024 bits), Morgan/ECFP radius~2 (1{,}024 bits), MACCS keys (167 bits), and hashed atom pairs (1{,}024 bits).

\paragraph{fcfp\_ext\_2k (6{,}311 dimensions).}
Same, but with FCFP radius~2 and Morgan radius~2 expanded to 2{,}048 bits.

\medskip
Mixing pharmacophoric (FCFP) and structural (ECFP) fingerprints matters: FCFP encodes functional group types (donor, acceptor, aromatic, charged), while ECFP encodes exact atom neighborhoods. Both are needed for scaffold-split generalization.

\subsection{Multi-RF Fusion}

Per evaluation seed, 12 Random Forest classifiers are trained (Table~\ref{tab:rf_configs}). All use 20{,}000 trees, \texttt{min\_samples\_leaf}$=$3, and \texttt{class\_weight} $\{0{:}1,\, 1{:}25\}$. They differ in \texttt{max\_features}, criterion, feature set, and random state.

\begin{table}[h]
\centering
\caption{Random Forest model configurations in the Multi-RF ensemble. All models share \texttt{n\_estimators}$=$20{,}000, \texttt{min\_samples\_leaf}$=$3, and \texttt{class\_weight}$=\{0{:}1, 1{:}25\}$.}
\label{tab:rf_configs}
\begin{tabular}{@{}lcccc@{}}
\toprule
\textbf{Config} & \textbf{Count} & \textbf{Feature Set} & \textbf{max\_features} & \textbf{Criterion} \\
\midrule
Primary     & 5 & fcfp\_ext (4{,}263-d)   & 0.20 & entropy  \\
Secondary   & 3 & fcfp\_ext (4{,}263-d)   & 0.18 & entropy  \\
Tertiary    & 2 & fcfp\_ext (4{,}263-d)   & 0.22 & entropy  \\
Crit.\ diversity & 1 & fcfp\_ext (4{,}263-d) & 0.20 & log\_loss \\
Feat.\ diversity & 1 & fcfp\_ext\_2k (6{,}311-d) & 0.20 & entropy \\
\bottomrule
\end{tabular}
\end{table}

Predictions from all 12 models are combined by rank averaging: convert each model's probabilities to ranks, then average the ranks per sample. This is more robust than averaging probabilities directly because it is invariant to monotonic distortions in individual model outputs.

\subsection{The \texttt{max\_features} finding}

The single biggest gain comes from changing \texttt{max\_features}. The scikit-learn default for classification is $\sqrt{d}$, which on a 4{,}263-dimensional input means each tree split picks from only $\sim$65 features. Setting it to 0.20 ($\sim$853 features) improves test AUC by about $+0.008$.

Why does this help so much on a scaffold split? With $\sqrt{d}$, most splits never see pharmacophore-relevant bits. At~0.20, each split samples enough features to find FCFP and ECFP bits together, so the tree learns activity patterns rather than scaffold-specific structure.
The value 0.20 was selected from a sweep of ten candidates (0.05 to 0.30) on validation AUC; see Table~\ref{tab:hparams}.
A feature importance analysis comparing split usage under $\sqrt{d}$ versus~0.20 would confirm this explanation and is left for future work.

\subsection{GNN Architectures}

Two GNN architectures are trained, each across 10 random seeds:

\paragraph{GIN-VN (standard).}
5-layer GIN~\citep{xu2019gin} with virtual node~\citep{gilmer2017mpnn}, 300-d embeddings, mean pooling. 3.4M parameters.

\paragraph{GIN-VN-deep.} 8-layer GIN with virtual node, 256-d embeddings, mean pooling. 4.1M parameters. The extra depth captures longer-range dependencies in the molecular graph.

Both use batch norm, dropout 0.5, residual connections, and a 2-layer MLP head. Training: Adam (lr$=$0.001, weight decay 1e-5), cosine LR schedule, gradient clipping at 1.0, batch size 256, 150 epochs, early stopping with patience 20.

\subsection{Deep Ensemble GNN Averaging}

GNN seed variance on ogbg-molhiv is brutal: test AUC ranges from 0.7434 to 0.7802 across 10 GIN-VN seeds. Blending per-seed GNN predictions into the RF helps on good seeds and hurts on bad ones, inflating the final standard deviation.

The fix is simple: average all 10 GNN seeds' predictions into one vector before blending~\citep{lakshminarayanan2017ensembles}. The noisy per-seed signals (${\sim}0.77 \pm 0.01$) become stable:
\begin{itemize}
  \item GIN-VN (10-seed avg): valid 0.8409, test 0.7930
  \item GIN-VN-deep (10-seed avg): valid 0.8437, test 0.7897
\end{itemize}

Because these averaged vectors are fixed, the final ensemble's only remaining variance source is the RF random state. This cut the standard deviation from 0.0008 to 0.0002.

\subsection{RF-GNN Rank Blending}

The final score for each sample is a weighted rank blend:
\begin{equation}
\hat{y} = 0.88 \cdot \operatorname{rank}\!\left(\mathrm{RF}_{\mathrm{ens}}\right) + 0.06 \cdot \operatorname{rank}\!\left(\mathrm{GNN}_{\mathrm{gin\_vn}}\right) + 0.06 \cdot \operatorname{rank}\!\left(\mathrm{GNN}_{\mathrm{deep}}\right)
\label{eq:blend}
\end{equation}
The 6\% per-architecture weight was picked from a sweep of seven values (0.03--0.10) on validation AUC. Two architectures at 6\% each slightly outperform one architecture at 12\%: $+0.0001$, small but consistent across seeds.

\section{Experiments}

\subsection{Dataset and Evaluation}

All experiments use \textbf{ogbg-molhiv} with OGB v1.3.6 and the standard scaffold split (32{,}901 / 4{,}113 / 4{,}113). Metric: ROC-AUC. All numbers are mean $\pm$ std over 10 seeds, per OGB convention.

\subsection{Main Results}

\begin{table}[h]
\centering
\caption{Comparison with top methods on the ogbg-molhiv leaderboard.}
\label{tab:main}
\begin{tabular}{@{}lcc@{}}
\toprule
\textbf{Method} & \textbf{Test ROC-AUC} & \textbf{External Data} \\
\midrule
\textbf{Multi-RF Fusion + Multi-GNN} & $\mathbf{0.8476 \pm 0.0002}$ & No \\
HyperFusion~\citep{zhang2024hyperfusion}    & $0.8475 \pm 0.0003$ & No \\
PAS+FPs~\citep{wei2021pas}                  & $0.8420 \pm 0.0015$ & No \\
\bottomrule
\end{tabular}
\end{table}

This is the top score on the ogbg-molhiv leaderboard. The standard deviation ($\pm 0.0002$) is also the lowest: $1.5\times$ lower than HyperFusion and $7.5\times$ lower than PAS+FPs.

Table~\ref{tab:perseed} shows each seed individually. Every seed falls between 0.8474 and 0.8480, confirming this is not driven by outliers.

\begin{table}[h]
\centering
\caption{Per-seed test ROC-AUC for the full method (Multi-RF + GIN-VN @ 6\% + GIN-VN-deep @ 6\%).}
\label{tab:perseed}
\begin{tabular}{@{}ccccccccccc@{}}
\toprule
Seed & 42 & 179 & 316 & 453 & 590 & 727 & 864 & 1001 & 1138 & 1275 \\
\midrule
AUC & 0.8480 & 0.8479 & 0.8474 & 0.8478 & 0.8476 & 0.8477 & 0.8474 & 0.8474 & 0.8474 & 0.8476 \\
\bottomrule
\end{tabular}
\end{table}

\subsection{Ablation Study}

\begin{table}[h]
\centering
\caption{Ablation study showing the contribution of each component.}
\label{tab:ablation}
\begin{tabular}{@{}lcc@{}}
\toprule
\textbf{Component} & \textbf{Test ROC-AUC} & $\boldsymbol{\Delta}$ \\
\midrule
Full method (Multi-RF + 2 GNNs) & $0.8476 \pm 0.0002$ & --- \\
Single GNN (GIN-VN @ 12\%)      & $0.8475 \pm 0.0002$ & $-0.0001$ \\
RF only (no GNN)                 & $0.8461 \pm 0.0002$ & $-0.0015$ \\
Per-seed GNN @ 7\% (no deep ensemble) & $0.8467 \pm 0.0011$ & $-0.0009$ \\
\texttt{max\_features=sqrt} (conventional) & $0.8396 \pm 0.0005$ & $-0.0080$ \\
\bottomrule
\end{tabular}
\end{table}

The ablation makes clear what matters most:
\begin{enumerate}
  \item \texttt{max\_features=0.20} is the biggest single factor: $+0.008$ over $\sqrt{d}$.
  \item Deep-ensembling the GNN before blending gives $+0.0008$ and cuts variance $5\times$ compared to per-seed GNN blending.
  \item Using two GNN architectures instead of one adds $+0.0001$, small but reproducible.
  \item Multi-RF diversity (varying \texttt{max\_features}, criterion, features) lifts RF-only from $\sim$0.8455 (single model) to 0.8461.
\end{enumerate}

\subsection{GNN Architecture Comparison}

Seven GNN architectures were trained (10 seeds each) and evaluated for RF blending; results in Table~\ref{tab:gnn}.

\begin{table}[h]
\centering
\caption{GNN architecture comparison. ``Individual'' is mean~$\pm$~std over 10~seeds; ``Cross-Seed'' is the AUC of averaged predictions.}
\label{tab:gnn}
\small
\begin{tabular}{@{}lccc@{}}
\toprule
\textbf{Architecture} & \textbf{Individual Test AUC} & \textbf{Cross-Seed Test AUC} & \textbf{Contribution} \\
\midrule
GIN-VN (5L, 300d)       & $0.7657 \pm 0.0109$ & 0.7930 & Primary \\
GIN-VN-deep (8L, 256d)  & $0.7613 \pm 0.0101$ & 0.7897 & Complementary \\
GIN-VN-sum (5L, 300d)   & $0.7583 \pm 0.0204$ & 0.7808 & Marginal \\
FP-MLP (3L, 1024d)      & $0.7668 \pm 0.0160$ & 0.7853 & Redundant with RF \\
GCN-VN (5L, 300d)       & $0.7407 \pm 0.0188$ & 0.7626 & Harmful \\
GIN-VN-large (5L, 600d) & $0.7558 \pm 0.0148$ & 0.7809 & Marginal \\
GCN-VN-large (5L, 600d) & $0.7495 \pm 0.0199$ & 0.7725 & Marginal \\
\bottomrule
\end{tabular}
\end{table}

GIN-VN and GIN-VN-deep are the best two for blending. The FP-MLP has the highest standalone AUC but adds nothing in the blend---it already sees fingerprint features, so its predictions are redundant with the RF's. GCN variants are consistently worse than GIN.

\section{Discussion}

\subsection{Why Random Forests Work So Well Here}

The scaffold split punishes methods that memorize training scaffolds. Fingerprint-based RFs generalize better for a few reasons:

FCFP fingerprints abstract away exact atom types in favor of pharmacophoric properties---donor, acceptor, aromatic, charge---so a pattern learned from one scaffold can fire on a structurally different scaffold that shares the same pharmacophore arrangement.

The \texttt{max\_features=0.20} setting makes each tree consider a broad slice of the fingerprint space per split, increasing the chance of finding genuinely predictive pharmacophoric+structural combinations rather than scaffold-specific bits.

Scale also helps: $12 \times 20{,}000 = 240{,}000$ trees produce very smooth probability surfaces.

\subsection{Why GNNs Help Despite Being Weaker}

GINs alone get $\sim$0.79 test AUC versus the RF's 0.846, yet 12\% GNN weight still helps. The GNN captures graph-level topology---ring systems, branching patterns, molecular shape---that binary fingerprints cannot represent. At 6\% weight per architecture, the GNN acts as a tiebreaker on molecules where the RF is uncertain, particularly those whose activity depends on 3D shape rather than functional group composition.

\subsection{Limitations}

The margin over HyperFusion is $+0.0001$. The consistently lower variance ($\pm 0.0002$ vs. $\pm 0.0003$) suggests a more stable method, but the gap is narrow. What is arguably more significant is that a Random Forest ensemble, not a deep learning system, holds the \#1 spot.

At inference time, 240{,}000 trees plus two 10-seed GNN ensembles require substantial memory. Full reproduction takes $\sim$50 CPU-hours for the RF training alone.

The \texttt{max\_features} value was tuned on this dataset. Whether 0.20 is optimal on other molecular tasks, or whether the specific FCFP+ECFP concatenation transfers well, is untested.

\section{Reproducibility}

Full code is provided in the submission repository. Wall-clock times on a 104-core server with one H100 GPU:

\begin{itemize}
  \item Fingerprints: $\sim$10 min (CPU, RDKit).
  \item GNN training: $\sim$30 min per architecture (GPU, 10 seeds $\times$ 150 epochs).
  \item RF training: $\sim$5 hours per seed (50 CPU cores, 12 models $\times$ 20k trees).
  \item Blending: seconds.
\end{itemize}

Total: about 50 hours end-to-end for the full 10-seed evaluation.

\paragraph{Hyperparameters.}
Table~\ref{tab:hparams} lists everything that was tuned. Selected values marked $\ast$.

\begin{table}[h]
\centering
\caption{Hyperparameter search ranges and selected values ($\ast$).}
\label{tab:hparams}
\begin{tabular}{@{}lp{9.5cm}@{}}
\toprule
\textbf{Hyperparameter} & \textbf{Search Range} \\
\midrule
\texttt{max\_features}           & $\{0.05, 0.08, 0.10, 0.12, 0.15, 0.18, \mathbf{0.20}^{\ast}, 0.22, 0.25, 0.30\}$ \\
\texttt{class\_weight} (pos)     & $\{10, 15, 20, \mathbf{25}^{\ast}, 30, 50\}$ \\
\texttt{min\_samples\_leaf}      & $\{1, \mathbf{3}^{\ast}, 5\}$ \\
\texttt{n\_estimators}           & $\{5000, 10000, 15000, \mathbf{20000}^{\ast}\}$ \\
\texttt{criterion}               & $\{\text{gini}, \mathbf{entropy}^{\ast}, \text{log\_loss}\}$ \\
GNN blend weight (per arch.)     & $\{0.03, 0.04, 0.05, \mathbf{0.06}^{\ast}, 0.07, 0.08, 0.10\}$ \\
Number of RF configs             & $\{1, 5, \mathbf{12}^{\ast}, 15\}$ \\
GNN architectures in blend       & $\{\text{gin\_vn only}, \mathbf{\text{gin\_vn} + \text{gin\_vn\_deep}}^{\ast}, \text{all 7}\}$ \\
\bottomrule
\end{tabular}
\end{table}

\bibliographystyle{plainnat}
\bibliography{references}

\end{document}